\documentclass[10pt,twocolumn,letterpaper]{article}

\usepackage{wacv}
\usepackage{times}
\usepackage{epsfig}
\usepackage{graphicx}
\usepackage{amsmath}
\usepackage{amssymb}
\usepackage{cite}

\def\wacvPaperID{610} %

\wacvfinalcopy %

\ifwacvfinal
\fi

\ifwacvfinal
\usepackage[breaklinks=true,bookmarks=false]{hyperref}
\else
\usepackage[pagebackref=true,breaklinks=true,colorlinks,bookmarks=false]{hyperref}
\fi

\begin{document}
\newcommand{\new}[1]{\textcolor{red}{#1}}
\newcommand{\val}[1]{\textcolor{blue}{[valentin: #1]}}
\newcommand{\arsha}[1]{\textcolor{blue}{[arsha: #1]}}
\newcommand{\ka}[1]{\textcolor{blue}{[karteek: #1]}}

\title{Masking Modalities for Cross-modal Video Retrieval}

\author{Valentin Gabeur$^{1,2}$ \hspace{4mm}
Arsha Nagrani$^{2}$ \hspace{4mm}
Chen Sun$^{2}$ \hspace{4mm}
Karteek Alahari$^{1}$ \hspace{4mm}
Cordelia Schmid$^{2}$ \\
$^1$ Inria$^*$  \hspace{20mm} $^2$ Google Research%
}

\maketitle
{\let\thefootnote\relax\footnote{$^*$ Univ.\ Grenoble Alpes, Inria, CNRS,
    Grenoble INP, LJK, 38000 Grenoble, France.}}

\begin{abstract}

Pre-training on large scale unlabelled datasets has shown impressive performance improvements in the fields of computer vision and natural language processing. Given the advent of large-scale instructional video datasets, a common strategy for pre-training video encoders is to use the accompanying speech as weak supervision. However, as speech is used to supervise the pre-training, it is never seen by the video encoder, which does not learn to process that modality. We address this drawback of current pre-training methods, which fail to exploit the rich cues in spoken language. Our proposal is to pre-train a video encoder using all the available video modalities as supervision, namely, appearance, sound, and transcribed speech. We mask an entire modality in the input and predict it using the other two modalities. This encourages each modality to collaborate with the others, and our video encoder learns to process appearance and audio as well as speech. We show the superior performance of our `modality masking' pre-training approach for video retrieval on the How2R, YouCook2 and Condensed Movies datasets.

\end{abstract}

\section{Introduction}
We live in a multimodal world, communicating through speech, visual signals and sound. This is reflected in the videos created and uploaded online---often they are accompanied by a highly informative audio track containing cues complementary to visual content.
Our goal in this work is to perform video retrieval with natural language queries.

\begin{figure}[t]
  \centering
  \includegraphics[width=\linewidth]{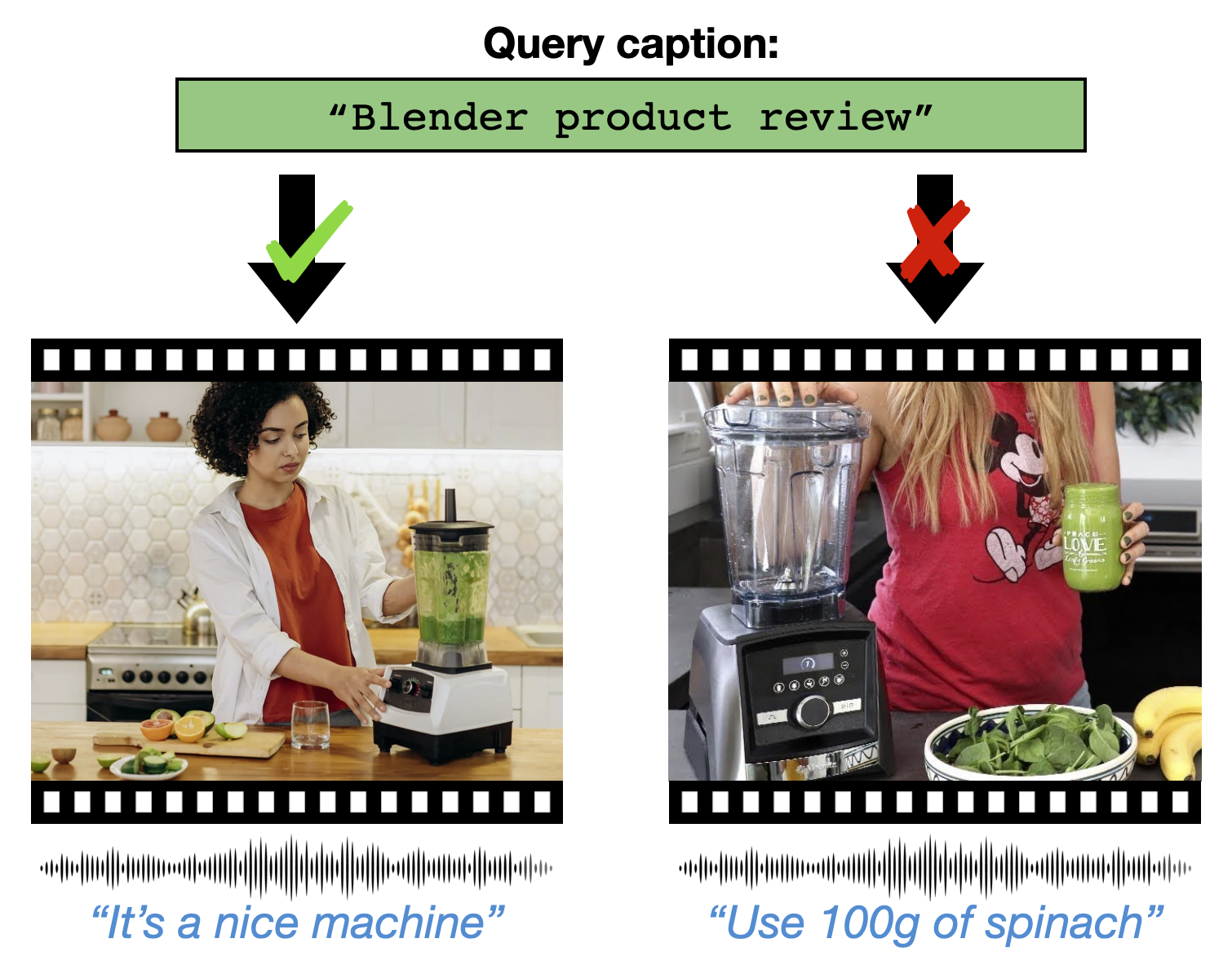}
  \caption{
Speech is part of the story! Video retrieval methods that focus on visual inputs alone are likely to miss out on key information (e.g., while both the examples above contain a blender, the speech (in blue) helps identify the one for a product review). In this work, we focus on learning a video encoder to effectively process RGB and audio features, as well as transcribed speech from instructional videos online, through a novel modality masking method. Our approach learns from unlabelled videos, without the need for expensive manual captions.
  }
\label{fig:asr_is_important}
\end{figure}

\begin{figure*}[htp]
  \centering
  \includegraphics[width=\textwidth]{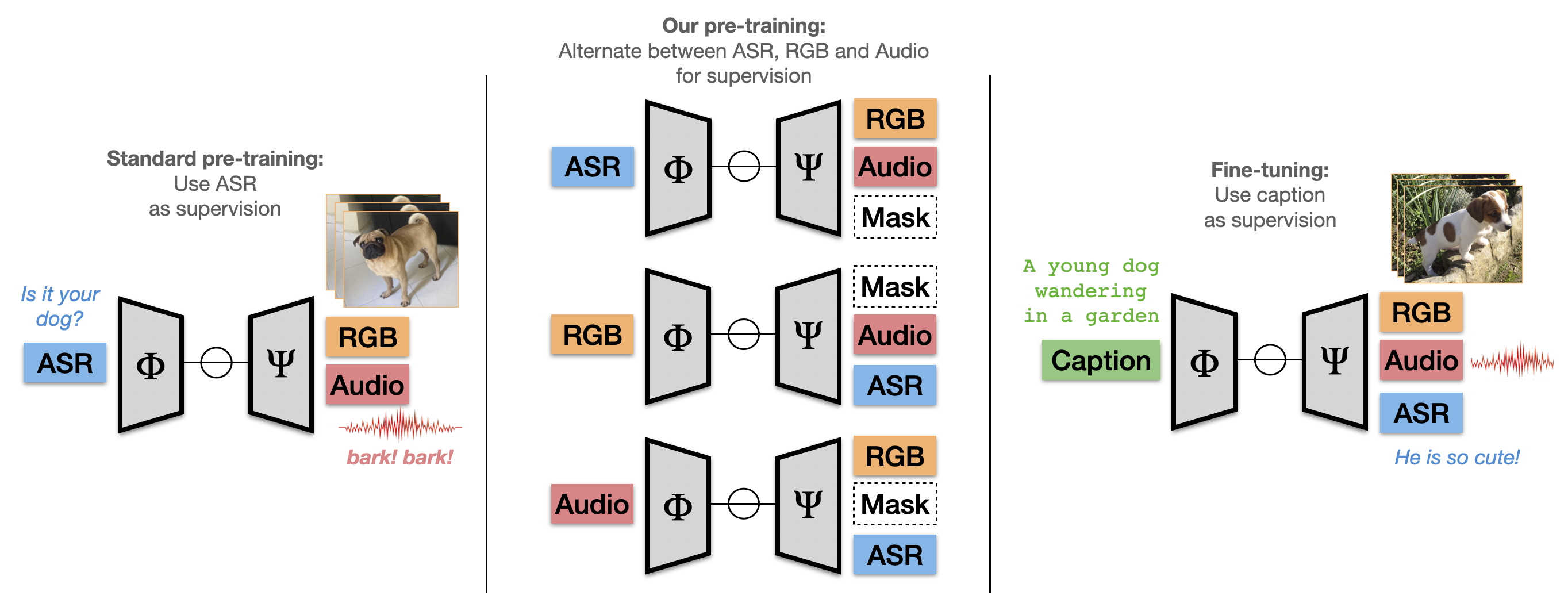}
  \caption{A common paradigm in learning from instructional videos is use transcribed speech (from ASR) (projected here using an encoder $\Phi$) to supervise a video encoder $\Psi$ (left). Instead, we train our video encoder $\Psi$ with three inputs -- RGB, audio and transcribed speech (ASR), and alternate between masking and predicting an entire modality at a time (middle). At the time of fine-tuning (right), our video encoder has been pre-trained to use all video modalities.}
  \label{fig:architecture}
\end{figure*}

While many popular video understanding works~\cite{xu2016msrvtt,krishna2017dense,miech2019MIL-NCE,patrick2021supportset,lei2021less,bain2021frozen} restrain the video signal to a sequence of visual frames, several approaches~\cite{liu2019use,miech2018learning,gabeur2020mmt} have progressed to incorporate information from different modalities through the use of pre-trained feature extractors called ``experts".
For videos, naturally composed of multimodal information, learning the optimal fusion of different modality `experts' is paramount. This challenge of multimodal learning is made more difficult by the scarcity of large manually-captioned video datasets. Existing datasets, e.g., \cite{xu2016msrvtt,li2020hero,Rohrbach2015LSMDC,krishna2017activitynet,Zhou2018YouCook2} remain small scale.
This has led to a several approaches utilising the large amount of instructional videos online~\cite{miech19howto100m, miech2019MIL-NCE, gabeur2020mmt, patrick2021supportset}, where transcribed speech (obtained with ASR) is closely linked to visual content, and hence a valuable source of supervision to train video encoders. Because of the proximity between text queries and speech, this approach presents the advantage of transferring well to text-to-video retrieval tasks. However, because the speech modality is used as a source of `pseudo' captioning labels, most of these works~\cite{miech19howto100m,miech2019MIL-NCE,gabeur2020mmt} only pre-train an encoder to process non-speech modalities (RGB, audio, etc), thereby not learning to combine speech and visual inputs effectively during pre-training. For many videos online, effectively processing speech is crucial for accurate video retrieval (Fig. \ref{fig:asr_is_important}).

In this work, we propose a novel pre-training strategy for learning multi-modal fusion from instructional videos (Fig. \ref{fig:architecture}, middle). We learn two encoders - the first being a video encoder ($\Psi$) that fuses experts from three modalities - RGB, transcribed speech (which we henceforth refer to as ASR for brevity), and audio. During pre-training we use a modality masking strategy, where we mask out an entire modality in the input of the video encoder, and try to predict an encoded version of this modality (encoded using a second encoder ($\Phi$)) from the other modalities. In this manner, the modality being predicted is effectively being used as `supervision' for the other two. At each batch, we mask a different modality, thereby learning a video encoder that is able to process all the modalities available in the video signal.

We make the following two contributions:
(i) We introduce a new pre-training approach for learning video representations that does not require costly manual annotations. Unlike previous works~\cite{miech2019MIL-NCE, gabeur2020mmt, li2020hero, miech19howto100m}, we train our video encoder with three inputs – RGB, audio and ASR, and alternate at each batch which one is used for supervision. At the time of fine-tuning, our video encoder has been pre-trained to use all video modalities.
(ii) We obtain competitive results on several standard text-to-video benchmarks.

\section{Related work}

\subsection{Multimodal Methods for Video}
Despite the fact that videos are often inherently multimodal, many popular works~\cite{xu2016msrvtt,krishna2017dense,miech2019MIL-NCE,patrick2021supportset,lei2021less,bain2021frozen} for video and language understanding discard the audio signal, potentially losing rich and varied additional information mentioned in speech or other background sounds. This may be due to the difficulty of jointly processing multiple modalities and the computational cost associated with processing such a high dimensional signal.
Another factor may also be an inherent bias in the annotation procedure for many datasets, for example, the LSMDC dataset~\cite{Rohrbach2015LSMDC} is annotated with audio descriptions (AD), which by nature must ignore the speech track, and in the ActivityNet dataset~\cite{krishna2017activitynet}, annotators were explicitly asked to ignore the audio track. The MSVD dataset~\cite{chen2011MSVD} videos simply do not have audio. 
In this work, we pretrain a video encoder using multiple video modalities, including audio and speech. We therefore also evaluate our approach on datasets such as How2R~\cite{li2020hero}, CMD~\cite{bain2020condensed} and YouCook2~\cite{Zhou2018YouCook2}, where speech plays an important role in understanding video content. 

\subsection{Experts for Video-to-Text Retrieval}
Because of the small scale of manually annotated text-to-video retrieval datasets as well as the high computational cost of processing pixels and raw audio signal directly, a popular approach for video retrieval has been to use pre-extracted features from `expert' models. These models are trained for diverse tasks and on multiple modalities such as face~\cite{parkhi2015deep}, scene~\cite{zhou2017places} and object recognition, action classification~\cite{carreira2017quo} and sound classification~\cite{gemmeke2017audio}. 
MoEE~\cite{miech2018learning}, CE~\cite{liu2019use} and MMT~\cite{gabeur2020mmt} all follow this paradigm, with the overall similarity for a video-text pair obtained as a weighted sum of each expert’s similarity with the text. 
In~\cite{ging2020coot}, the authors propose a two-branches architecture that models the interactions between different levels of granularity in both the visual modality and the text modality. 
More recently, several works~\cite{PortilloQuintero2021ASF,Dzabraev2021MDMMT} have shown the superiority of using the CLIP~\cite{Radford2021CLIP} model to extract appearance features, therefore leveraging the 400 million (image, caption) pairs it was trained on.

\subsection{Pre-training for Video and Language}
Since the release of the HowTo100M dataset~\cite{miech19howto100m}, a large instructional video dataset, there has been a renewed interest in leveraging large-scale pre-training to
improve video-text representations for tasks such as video question-answering~\cite{seo2021look, li2020hero}, text-to-video retrieval~\cite{miech19howto100m,patrick2021supportset, gabeur2020mmt}, action recognition~\cite{miech2019MIL-NCE, Sun2019cbt,nagrani2020speech2action,Alayrac2020Versatile} and video captioning~\cite{zhou2018end,huang2020multimodal}.
In NLP, BERT~\cite{devlin2018bert} and its variants have popularized the `masked language modelling' self-supervised technique for pre-training: wherein words in the input are randomly masked and the training objective is tasked with predicting their encodings. This technique has been extended to train visual and language encoders (eg. VideoBERT~\cite{Sun2019VideoBERT}, CBT~\cite{Sun2019cbt}, ViLBERT~\cite{lu2019vilbert}, Hero~\cite{li2020hero} etc). All such works, only mask a proportion of the input (usually 15\%).
Contrary to natural language, the visual signal and audio signal of a video are continuous and highly redundant. A masked video or audio segment can be easily estimated from its neighboring frames. To address this problem, we mask out an entire modality in the input, forcing our model to learn difficult cross-modal interactions. 
 
\section{Methodology}
In this section, we first describe the common  pre-training approach for learning from instructional videos, where the ASR is used to supervise a visual encoder. We then present our strategy to pre-train a video encoder $\Psi$ on three video modalities: RGB, Audio and ASR, by using each of them to supervise the others in an alternating manner. After pre-training, our video encoder has learnt to attend across all modalities in a video, and can be fine-tuned on video-text datasets for the task of video retrieval.

\subsection{Standard Pre-Training}
As a video representation learning pre-training strategy, several previous works~\cite{miech19howto100m, miech2019MIL-NCE, gabeur2020mmt} use the speech modality as supervision to train a video encoder on the other video modalities. Illustrated on the left side of Fig.~\ref{fig:architecture}, this approach involves the estimation of a speech representation by a query encoder $\Phi$ and a video representation by a video encoder $\Psi$. The training objective is usually a standard metric learning objective (maximising the similarity between the speech representation and the video representation if they are extracted from the same video, minimizing the similarity between randomly selected speech and video). At the time of fine-tuning (right side of Fig.~\ref{fig:architecture}, the query encoder $\Phi$ is used to encode the caption while the video encoder $\Psi$ is processing all video modalities, including speech.

The main drawback of this approach is that the video encoder is not pre-trained on speech since that modality is used as pre-training supervision. At the end of pre-training, the video encoder has hence been denied the opportunity to learn complex cross-modal interactions between RGB and speech. The video encoder only learns to process speech during fine-tuning. This is a major limitation as speech may be an integral part of the video signal and encode crucial information for video retrieval.

\subsection{Alternating Modality-Masking Pre-Training}
We propose a new approach for pre-training a video encoder on a large-scale dataset of raw videos like HowTo100M~\cite{miech19howto100m}, which does not contain captioning labels.
In order for our video encoder to be pre-trained on all video modalities, \textbf{including ASR}, we propose to not only use ASR supervision, but to alternate between three objectives (Middle section of Fig.\ref{fig:architecture}):

\begin{enumerate}
  \item Use ASR as supervision to train the video encoder $\Psi$ on processing RGB + Audio as inputs 
  \item Use RGB as supervision to train the video encoder $\Psi$ on processing Audio + ASR as inputs
  \item Use Audio as supervision to train the video encoder $\Psi$ on processing RGB + ASR as inputs
\end{enumerate}

At each training batch, we randomly pick one of those objectives. We therefore randomly pick a modality in $\{RGB, Audio, ASR\}$ to serve as the supervising modality processed by the query encoder, while the other two modalities act as the collaborating modalities processed by the video encoder.
Let us take the example of a training batch for which RGB has been selected as the supervising modality. For each video of the batch, its sequence of RGB features will be processed by the query encoder $\Phi$ to obtain a query representation. The features of the other two modalities of the video, in this case Audio and ASR, will be processed by the video encoder $\Psi$ to extract both cross-modal and temporal information and obtain a video representation. We will then proceed to optimize the parameters of our encoders so that the query and video representations of a same video are similar while the representations of different videos in the batch are dissimilar.

More formally, for each video clip $v_i$ in the training batch, we separate the expert features in two sets: $q_i$ are the features obtained from the supervising modality and $c_i$ are the features obtained from the collaborating modalities. We then use our query encoder $\Phi$ to compute a representation $\Phi(q_i)$ of the supervising modality. Similarly, our video encoder $\Psi$ will compute a representation $\Psi(c_i)$ of the collaborating modalities. 

During fine-tuning (right side of Fig.\ref{fig:architecture}), our video encoder $\Psi$ is provided with all the modalities present in the video signal, all of which it has seen before during pre-training, and has hence acquired the ability to model cross-modal complex correlations.

Although our video encoder $\Psi$ only ever receives two modalities at a time during pre-training, they are different at each batch. We therefore need a video encoder capable of processing the three video modalities, but at each batch, one of them (the supervising modality) is "masked out", it is simply not provided to $\Psi$. In the next section we describe the architecture of that encoder.

\subsection{The Multi-Modal Transformer}
For our video encoder $\Psi$, we use the Multi-Modal Transformer described in~\cite{gabeur2020mmt}. It consists in a Transformer encoder that is fed features from different video modalities. The self-attention mechanism of the Transformer allows each token to attend to all the others, therefore being able to process information across both time and modalities. The choice of the MMT architecture for our modality-masking pre-training approach is justified by its capacity to elegantly handle missing modalities. In fact, all the transformer layers parameters are shared across all input features, and therefore modalities. That means that even if one modality is masked from the input of MMT, the parameters of all layers will still be optimized. All parameters needed for the downstream task are optimized at each batch, independently of the chosen objective.
This is in contrast to the MoEE style of architecture where there is a dedicated encoding branch for each modality. In the case of a missing expert stream, zeros will be fed, thereby wasting computation for that whole branch.

\subsection{Loss Function}
For both pre-training and fine-tuning, we optimize both query encoder $\Phi$ and video encoder $\Psi$ to provide similar representations when their input features come from the same video clip and dissimilar representations when they come from different clips.
We train our model with the bi-directional max-margin ranking loss~\cite{Karpathy2014DeepFE}:
\begin{equation} \label{eq:loss}
\mathcal{L} = \frac{1}{B}\sum_{i=1}^{B} \sum_{j \neq i} \Big[ \max(0, s_{ij} - s_{ii} + m) + \max(0, s_{ji} - s_{ii} + m)\Big],
\end{equation}
where $B$ is the batch size, $s_{ij} = s(\Phi(q_i),\Psi(c_j))$, the similarity score between query representation $\Phi(q_i)$ of video $v_i$ and video representation $\Psi(c_j)$ of video $v_j$, and $m$ is the margin. This loss enforces the similarity between true representation pairs $s_{ii}$ to be higher than the similarity between negative samples $s_{ij}$ or $s_{ji}$, for all $i \neq j$, by at least $m$.
This will have the effect of gathering similar captions and videos together in the embedding space, thereby allowing video retrieval to be performed by ranking videos according to their proximity with the query.

\subsection{Selection of Modalities}
We choose the modalities RGB, Audio and ASR in this work, largely because they often represent complementary aspect of the video signal. Although audio and speech are both extracted from the audio signal, the expert models extracting features for those modalities have been pre-trained on different tasks and present specialized architectures dedicated to those tasks. The CNN used to encode the audio (sounds) features is not capable to extract spoken language nor has it been trained on that task.

\section{Experiments}
We first describe the text-video datasets that our model is trained and evaluated on (sec. \ref{subsec:datasets}), then present implementation details (sec. \ref{subsec:impl}) and ablation studies (sec. \ref{subsec:ablation}). Finally, we compare to the state-of-the-art for video retrieval (sec. \ref{subsec:sota}).

\subsection{Datasets and Metrics}
\label{subsec:datasets}
Since the focus of our work is the effective encoding of ASR, audio and visual information, we evaluate on video datasets that contain multimodal captions (i.e., captions that refer to the content in the speech as well). For each dataset, we manually inspect 100 caption-video pairs at random to determine the percentage of captions that are related to what is being said in the video. %
For example, the caption ``Someone talking about love" requires knowledge of the speech in the video, whereas ``A woman with a red dress" does not. 
Results are reported below.\\
\noindent\textbf{HowTo100M~\cite{miech19howto100m}} is a very large-scale dataset of over 1M YouTube instructional videos that amounts to about 15 years of video. This dataset was not manually annotated with captions, but is a valuable source of data for self-supervised learning because of the high correlation between visual, audio and speech information in its videos. We only use this dataset to pre-train our model.\\
\noindent\textbf{How2R~\cite{li2020hero}} features 47,369 clips extracted from the HowTo100M dataset videos and split into a training, validating and testing set. The clips are 17s long on average, and annotated with a caption. Our manual inspection of 100 captions yields 54 captions related to speech. Because it has the same domain as our pre-training dataset, we run ablation studies on this dataset. \\
\noindent\textbf{MSR-VTT~\cite{xu2016msrvtt}} contains 10K YouTube videos with 200K descriptions. Following other works~\cite{liu2019use}, we train on 9K train+val videos and report results on the 1K-A test set. After manual inspection, we find that 12\% of the captions are related to speech.\\
\noindent\textbf{Condensed Movies Dataset (CMD) ~\cite{bain2020condensed}} consists of 33,976 clips extracted from 3,605 movies. Our manual verification process indicates that approximately 60\% of CMD descriptions are related to speech.\\
\noindent\textbf{YouCook2 ~\cite{Zhou2018YouCook2}} consists of 176 hours of cooking videos. The videos are segmented into 13,829 clips, each annotated with a sentence describing a step of the recipe.
We follow~\cite{miech19howto100m} and evaluate our model on 3,350 clips that are not present in HowTo100M. We found about 70\% of YouCook2 captions are related to the speech in the video. For example, in the case of a video annotated with ``add corn starch", paying attention to the speech: ``I now use corn starch" is a strong cue indicating that we are not dealing with flour. \\
\noindent\textbf{ActivityNet captions ~\cite{krishna2017activitynet}} consists of 20K YouTube videos annotated with several sentences. We follow~\cite{zhang2018HSE} and concatenate the sentences to obtain a paragraph annotation for each video. We found that only 1\% of the captions relate to speech. The authors of this dataset informed us that the annotators were explicitly asked to ignore the audio. It was turned off by default for the annotation process.\\
\noindent\textbf{Metrics.} We evaluate the performance of our approach on the following standard retrieval metrics: recall at rank $N$ (R@$N$, higher is better), median rank (MdR, lower is better) and mean rank (MnR, lower is better). For each metric we run the experiment with 3 random seeds and report the mean and standard deviation. We report the test-set performances of our model for the epoch where the validation-set geometric mean of R@1, R@5 and R@10 is maximal.  

\subsection{Implementation Details}
\label{subsec:impl}
\noindent\textbf{Pre-trained experts.} Both encoders use pre-trained expert models for extracting features from each video modality. We use the following 3 experts: \\
\noindent\textbf{RGB} features are extracted from S3D~\cite{Xie2017S3D} trained on the Kinetics action recognition dataset. We extract one RGB feature of dimension 1024 per second of video.\\
\noindent\textbf{Audio} features are extracted using VGGish model~\cite{Hershey2017VGGish} trained on the YouTube8M dataset~\cite{Abu2016YT8M}. We extract one audio feature of dimension 128 per second of video.\\
\noindent\textbf{ASR} transcripts are obtained from the closed captions accompanying videos on YouTube. Words are encoded with BERT-base-cased~\cite{devlin2018bert}. We obtain one speech feature of dimension 768 for each wordpiece from the ASR.

\noindent\textbf{Query encoder $\Phi$ for pre-training.}
In the case when the masked modality is either RGB or audio, we encode it using the Multi-Modal-Transformer (MMT)~\cite{gabeur2020mmt} model. We follow~\cite{gabeur2020mmt} and use a 4 layer, 4 head version of MMT. 
For any modality presented to the query encoder, we do not encode input sequence of features with temporal embeddings. We found that this made the pre-training objective trivially easy to solve - we hypothesise that this is because temporal information allows the encoders to align silences in the ASR and audio features (as both are extracted from the same audiotrack). For example, both encoders would be able to determine the presence and absence of speech from the ASR and audio modalities. The similarity can then be maximised based on this temporal alignment, instead of on the video semantics, leading to performance drops. 
In the case when the masked modality is ASR, we follow~\cite{gabeur2020mmt} and process the speech words with a pre-trained BERT model. For memory constraints, we limit BERT input to 30 consecutive wordpieces, randomly sampled from the ASR. The representation extracted from the BERT [CLS] token is projected by 3 different gated embedding units (one for each modality) to obtain our query representation $\Phi(q_i)$. \\

\noindent\textbf{Query encoder $\Phi$ for fine-tuning.}
During fine-tuning, we use the captions as supervision. For fine-tuning on MSRVTT, YouCook2 and ActivityNet, we follow the procedure introduced in MMT~\cite{gabeur2020mmt}: we process the caption with the Bert-based query encoder that we pre-trained earlier. We limit BERT input to 30 consecutive wordpieces, randomly sampled in the caption.
On the How2R and CMD datasets, we found out that using the pre-trained Bert-based query encoder for encoding the captions resulted in rapid over-fitting -- on the other hand, freezing the weights in the query encoder lead to poor performance. We therefore follow the approach outlined in MoEE~\cite{miech2018learning} and use a net-VLAD layer to aggregate the caption word embeddings (obtained by a frozen BERT model), to obtain the final caption representation. The caption representation is then projected by 3 different gated embedding units. \\

\noindent\textbf{Video encoder $\Psi$.}
This is implemented using the Multi-Modal Transformer(MMT)~\cite{gabeur2020mmt} as our video encoder. We use a 4 layers x 4 heads version of MMT with a dropout probability of 10\%, a hidden size $d_{model}$ of 512, and an intermediate size of 3072. We initialize the aggregated embeddings of MMT with a max-pooling aggregation of the modality features. In the case of a masked modality (pre-training) or when a modality is not available in the video, no features are provided to MMT and the aggregated embedding for that modality becomes a zero vector. 
For all our experiments, we only use the sequences of features extracted by our RGB, audio and ASR pre-trained experts. The parameters of those feature extractors are kept frozen. For memory constraints, we provide the video encoder with sequences of maximum 30 features for the RGB and audio modalities, and maximum 128 features for the ASR. In case more features are available in the video, they are randomly sampled. \\

\noindent\textbf{Hyperparameters.}
For each dataset, we estimate the hyperparameters by running a grid search on the corresponding validation set. We use the Adam optimizer for all the experiments. 

For pre-training on HowTo100M, we use a batch size of 1,200 videos, an initial learning rate of 1e-4, which we decay by a 0.98 multiplicative factor every 2K optimisation steps, and train for 400K steps. 
We randomly crop HowTo100M videos into segments of 30 seconds.

For training from scratch or fine-tuning on MSR-VTT or YouCook2, we use a batch size of 32 videos, an initial learning rate of 5e-5, which we decay by a 0.95 multiplicative factor every 1K optimisation steps, and train for 50K steps.
For training from scratch or fine-tuning on CMD or How2R, we use an initial learning rate of 5e-5, which we decay by a 0.90 multiplicative factor every 375 optimisation steps, and train for 20K steps. We use a batch size of 32 videos on How2R and 64 videos on CMD.
For training from scratch or fine-tuning on ActivityNet, we use a batch size of 24 videos, an initial learning rate of 5e-5, which we decay by a 0.90 multiplicative factor every 1K optimisation steps, and train for 50K steps.
For training on HowTo100M, MSR-VTT, YouCook2 or ActivityNet, the bidirectional max-margin ranking loss margin is set to 0.05. For training on How2R or CMD, it is set to 0.2.
\\

\noindent\textbf{Running time.}
Pre-training our model on HowTo100M takes 12 days on 8 V-100 GPUs. Fine-tuning on MSRVTT, How2R, CMD, YouCook2 or ActivityNet takes about 4 hours on a single V-100 GPU.

\subsection{Ablation Analysis}\label{subsec:ablation}
We perform three ablation studies to: (i) show the effect of varying the masking probability of ASR, $p$, during pre-training; (ii) demonstrate the need of complete modality masking over partial modality masking; and (iii) compare multi-modal retrieval results to those with a single modality. \\

\noindent\textbf{Effect of the ASR masking probability $p$.}
Table~\ref{table:settings_probs} shows the impact of different masking probabilities during pre-training on HowTo100M. The probability $p$ refers to the probability of masking our ASR and feeding in only RGB and audio to the candidate encoder (this is the common pre-training paradigm, where ASR is effectively `supervising' our video encoder). The rest of the time is equally split between masking out audio and RGB. Hence if $p=0.8$, we mask out ASR 80\% of the time, RGB 10\% of the time, and audio 10\% of the time. Note that this is equivalent to weighting the loss (Eq.~\ref{eq:loss}) differently depending on which modality is masked. We report results on the validation set of How2R after fine-tuning on the training set of How2R. We show that the common pre-training paradigm of always using the ASR to supervise RGB and audio ($p=1.0$, first line) does not provide the best results. It is better to also use audio and RGB as supervision in order to pre-train the video encoder on speech. For the rest of the experiments, we set $p=0.8$ during pre-training.

\begin{table}[t]
\begin{center}
\caption{The effect of the masking probability for transcribed speech (ASR) $p$, where $p=1.00$ refers to the case where ASR is masked 100\% of the time, and predicted from audio and RGB. Results are reported on the validation set of How2R after fine-tuning. We note that performance improves when $p<1$, but remains relatively robust to different values.}
\label{table:settings_probs}
\small
\begin{tabular}{l | c c c c c}
\multicolumn{1}{c}{} & \multicolumn{5}{c}{Text $\implies$ Video} \\
$p$ & R@1$\uparrow$ & R@5$\uparrow$ & R@10$\uparrow$ & MdR$\downarrow$ & MnR$\downarrow$ \\
\hline
1.00 & $3.1_{\!\pm\!\text{0.1}}$ & $9.9_{\!\pm\!\text{0.3}}$ & $15.5_{\!\pm\!\text{0.3}}$ & $97.0_{\!\pm\!\text{2.2}}$ & $292.3_{\!\pm\!\text{4.7}}$ \\
0.90 & $2.9_{\!\pm\!\text{0.0}}$ & $9.5_{\!\pm\!\text{0.0}}$ & $15.2_{\!\pm\!\text{0.0}}$ & $96.5_{\!\pm\!\text{0.0}}$ & $271.9_{\!\pm\!\text{0.0}}$ \\
0.80 & $\textbf{3.7}_{\!\pm\!\text{0.1}}$ & $\textbf{11.5}_{\!\pm\!\text{0.1}}$ & $\textbf{17.8}_{\!\pm\!\text{0.3}}$ & $\textbf{79.0}_{\!\pm\!\text{0.0}}$ & $270.7_{\!\pm\!\text{2.5}}$ \\
0.70 & $3.5_{\!\pm\!\text{0.1}}$ & $\textbf{11.5}_{\!\pm\!\text{0.2}}$ & $17.6_{\!\pm\!\text{0.1}}$ & $80.7_{\!\pm\!\text{0.9}}$ & $\textbf{267.8}_{\!\pm\!\text{1.4}}$ \\
0.33 & $3.5_{\!\pm\!\text{0.2}}$ & $\textbf{11.5}_{\!\pm\!\text{0.3}}$ & $\textbf{17.8}_{\!\pm\!\text{0.2}}$ & $82.0_{\!\pm\!\text{1.4}}$ & $269.8_{\!\pm\!\text{1.5}}$ \\
\end{tabular}
\end{center}
\end{table}

\noindent\textbf{Advantage of complete modality masking over partial masking.} 
Several recent works~\cite{Sun2019VideoBERT,Sun2019cbt,lu2019vilbert,li2020hero} pre-train a video encoder by partially masking a modality (eg: masking 15\% of video frames). We instead mask 100\% of the modality. We have compared our approach with masking 15\%, 50\% or 85\% of the supervising modality tokens. To not make the task trivial we did not provide the query encoder $\Phi$ with 100\% of the supervising modality tokens, but only with the feature tokens that were masked from the video encoder $\Psi$. We used the setting which obtained best results for 100\% masking, i.e., ASR is used as supervision for 80\% of the batches, audio 10\% and RGB 10\%.
 
Recall@10 results on the validation set of How2R are:
From scratch (no pre-training): 12.9,
Masking 15\%: 16.1,
Masking 50\%: 16.2,
Masking 85\%: 16.8,
Masking 100\%: 17.8.
 
The results for the partial masking pre-training show a lower performance compared to 100\% masking. We also noticed that during pre-training the loss for the partial masking experiments was lower than the loss for the 100\% masking experiment. This can be attributed to the fact that the query encoder $\Phi$ and video encoder $\Psi$ are both provided with some of the supervising modality features, making the pre-training task easier, and therefore less effective. This is particularly the case for the audio and visual modality because of their continuity and high redundancy.\\

\begin{figure}[h]
  \centering
  \includegraphics[width=\linewidth]{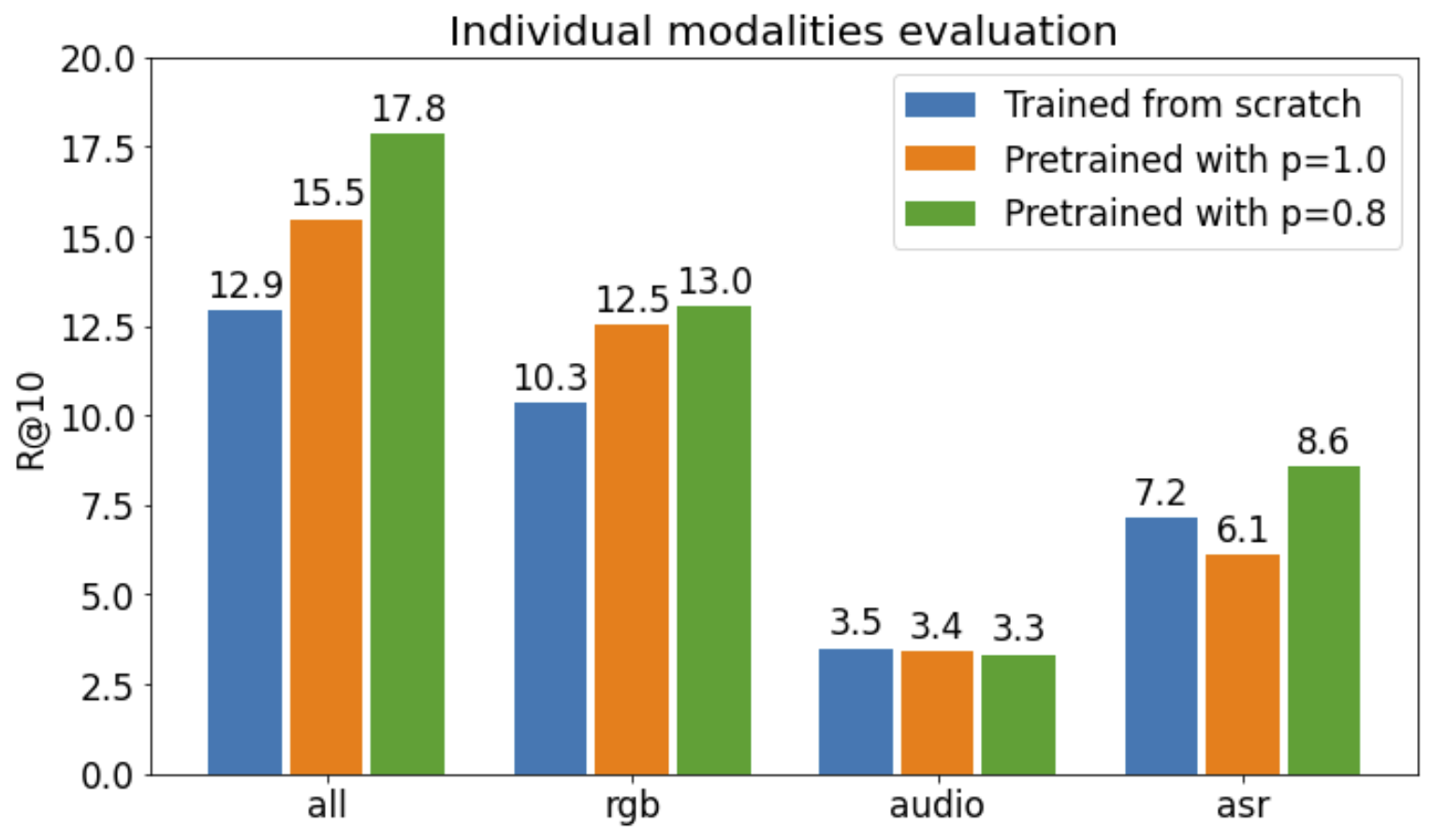}
  \caption{Impact of the pre-training approach on the retrieval of a single modality. We report results on the val set of How2R using R@10. (Best viewed in colour.)}
\label{fig:indiv_mods}
\end{figure}

\noindent\textbf{Impact of pre-training on single-modality retrieval.}
We further evaluate our pre-training approach by fine-tuning the model on a single video modality. In this case, for each video in the How2R validation set, our video encoder is only provided with the features of one modality, either RGB, Audio or ASR. We report the results using R@10 in Fig.~\ref{fig:indiv_mods}. When only pre-training with ASR supervision ($p=1.0$, orange), our video encoder only processes RGB and audio inputs. In this case, we note that pre-training helps when fine-tuning only on the RGB modality, but not on the audio modality. As expected, this setting leads to a performance drop on ASR, as the video encoder has never seen ASR inputs during pre-training.
This is not the case for our alternating modality masking approach where the video encoder was sometimes provided with ASR features and therefore learns to process that modality. We note that our alternating masking approach ($p=0.8$, green) provides improvements overall, as well as for each modality independently (other than for audio which does not seem to benefit from pre-training).

\begin{figure*}[h!]
  \centering
  \includegraphics[width=\textwidth]{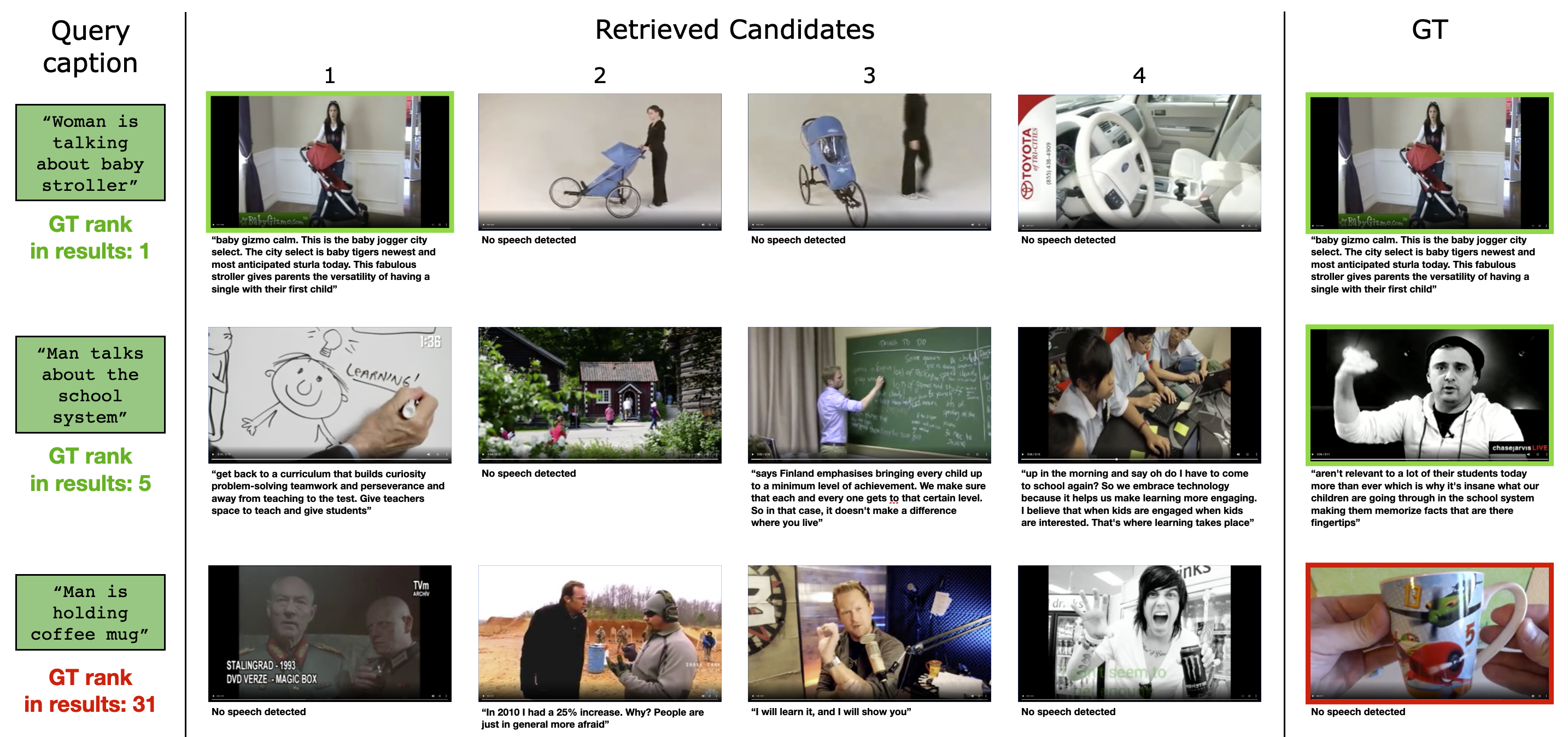}
  \caption{Qualitative results of our retrieval method on the MSR-VTT dataset. For each query, we show frames and ASR from the top 4 ranked videos as well as for the ground truth video. We indicate the rank of the ground-truth video in our retrieval results (highlighted in green when it is in the top-5 retrieved results, or red otherwise) on the left under the query.  Note that there are 1000 candidate videos in the test set. (Best viewed on screen.)}
  \label{fig:qualitative_results}
\end{figure*}

\subsection{Comparison to the State of the Art}\label{subsec:sota}

Results on How2R are provided in Table \ref{table:How2R}. The original paper introducing the How2R dataset~\cite{li2020hero} tackles the task of moment localization in a video clip. We re-purpose the How2R dataset for the task of video retrieval where each moment and its description are considered as a different video-caption pair.
We reproduce the MoEE approach~\cite{miech2018learning} on this dataset, and show that our method trained from scratch significantly outperforms MoEE. We also implement the MMT pre-training approach~\cite{gabeur2020mmt} (equivalent to p=1.0) with our features, and compare it with our modality masking pre-training (p=0.8) approach. The large performance improvement obtained with our approach demonstrates the advantage of pre-training the video encoder on speech before fine-tuning on the How2R dataset, which has more than half of its captions related to speech.

\begin{table}[h!]
\begin{center}
\caption{Text to Video retrieval results on the How2R~\cite{li2020hero} benchmark. $\dagger$ Our implementation on this dataset using only our RGB, audio and ASR features. sc: trained from scratch on How2R. pt: pre-trained on the HowTo100M dataset, then fine-tuned on How2R.}
\label{table:How2R}
\scriptsize
\begin{tabular}{l | c c c c c}
\multicolumn{1}{c}{} & \multicolumn{5}{c}{Text $\implies$ Video} \\
Method & R@1$\uparrow$ & R@5$\uparrow$ & R@10$\uparrow$ & MdR$\downarrow$ & MnR$\downarrow$ \\
\hline
Random & 0.0 & 0.1 & 0.2 & 2009.5 & 2009.5 \\
MoEE (sc)~\cite{miech2018learning}$\dagger$ & 2.2$_{\!\pm\!\text{0.1}}$ & 7.8$_{\!\pm\!\text{0.1}}$ & 12.9$_{\!\pm\!\text{0.3}}$ & 118.7$_{\!\pm\!\text{1.2}}$ & 389.5$_{\!\pm\!\text{1.8}}$ \\
Ours (sc) & 2.3$_{\!\pm\!\text{0.2}}$ & 8.3$_{\!\pm\!\text{0.3}}$ & 13.6$_{\!\pm\!\text{0.2}}$ & 106.0$_{\!\pm\!\text{2.2}}$ & 312.5$_{\!\pm\!\text{1.2}}$ \\
MMT (pt)~\cite{gabeur2020mmt}$\dagger$ & 2.9$_{\!\pm\!\text{0.0}}$ & 9.1$_{\!\pm\!\text{0.2}}$ & 14.5$_{\!\pm\!\text{0.2}}$ & 96.0$_{\!\pm\!\text{2.2}}$ & 314.3$_{\!\pm\!\text{1.7}}$ \\
Ours (pt p=0.8) & $\textbf{3.4}_{\!\pm\!\text{0.2}}$ & $\textbf{11.6}_{\!\pm\!\text{0.2}}$ & $\textbf{18.2}_{\!\pm\!\text{0.3}}$ & $\textbf{75.3}_{\!\pm\!\text{0.9}}$ & $\textbf{277.1}_{\!\pm\!\text{2.3}}$ \\
\end{tabular}
\end{center}
\end{table}

Results on CMD are provided in Table \ref{table:CMD}. Unlike the original CMD paper~\cite{bain2020condensed}, we remove actor names from the captions. We re-implement MoEE~\cite{miech2018learning} on this modified dataset using our features, and demonstrate that our pre-training approach provides a significant improvement in performance. Note that this is despite the large variation in domain between pre-training and fine-tuning -- while we pre-train on instructional videos from YouTube, CMD consists of short clips extracted from movies. 

\begin{table}[h!]
\begin{center}
\caption{Results on the Condensed Movies Dataset (CMD)~\cite{bain2020condensed}. $\dagger$ Our implementation on this dataset using only our RGB, audio and ASR features. $\ddagger$ Our implementation on this dataset using the code and all the features provided by the authors of CMD~\cite{bain2021frozen}.  sc: trained from scratch on CMD. pt: pre-trained on the HowTo100M dataset, then fine-tuned on CMD.}
\label{table:CMD}
\scriptsize
\begin{tabular}
{l | c c c c c}
\multicolumn{1}{c}{} & \multicolumn{5}{c}{Text $\implies$ Video} \\
Method & R@1$\uparrow$ & R@5$\uparrow$ & R@10$\uparrow$ & MdR$\downarrow$ & MnR$\downarrow$ \\
\hline
Random & 0.0 & 0.1 & 0.2 & 3284.5 & 3284.5 \\
MoEE (sc)~\cite{miech2018learning}$\dagger$ & 3.2$_{\!\pm\!\text{0.1}}$ & 9.9$_{\!\pm\!\text{0.3}}$ & 14.9$_{\!\pm\!\text{0.4}}$ & 142.7$_{\!\pm\!\text{0.5}}$ & 532.7$_{\!\pm\!\text{5.7}}$ \\
CMD (sc)~\cite{bain2020condensed}$\ddagger$ & 2.6 & 10.2 & 16.2 & 102 & 377.7 \\
Ours (sc) & 4.6$_{\!\pm\!\text{0.1}}$ & 13.5$_{\!\pm\!\text{0.2}}$ & 19.5$_{\!\pm\!\text{0.1}}$ & 89.7$_{\!\pm\!\text{1.2}}$ & 396.5$_{\!\pm\!\text{5.5}}$ \\
Ours (pt p=0.8) & $\textbf{5.8}_{\!\pm\!\text{0.2}}$ & $\textbf{15.8}_{\!\pm\!\text{0.2}}$ & $\textbf{22.4}_{\!\pm\!\text{0.1}}$ & $\textbf{73.7}_{\!\pm\!\text{1.7}}$ & $\textbf{369.6}_{\!\pm\!\text{4.6}}$ \\
\end{tabular}
\end{center}
\end{table}

Table \ref{table:YouCook2} presents results on YouCook2. Due to the high importance of the speech modality in this dataset, pre-training with our approach (pt p=0.8) yields considerable performance improvement, compared to the standard pre-training approach (MMT) that does not pre-train the video encoder on the speech modality.

\begin{table}[h!]
\begin{center}
\caption{Results on the YouCook2 dataset~\cite{Zhou2018YouCook2}. $\dagger$ Our implementation on this dataset. sc: trained from scratch on YouCook2. pt: pretrained on the HowTo100M dataset, then fine-tuned on YouCook2. }
\label{table:YouCook2}
\scriptsize
\begin{tabular}{l | c c c c c}
\multicolumn{1}{c}{} & \multicolumn{5}{c}{Text $\implies$ Video} \\
Method & R@1$\uparrow$ & R@5$\uparrow$ & R@10$\uparrow$ & MdR$\downarrow$ & MnR$\downarrow$ \\
\hline
Random & 0.03 & 0.15 & 0.3 & 1675 & 1675 \\
Ours (sc) & 16.6$_{\!\pm\!\text{0.2}}$ & 37.4$_{\!\pm\!\text{0.3}}$ & 48.3$_{\!\pm\!\text{0.1}}$ & 12.0$_{\!\pm\!\text{0.0}}$ & 95.5$_{\!\pm\!\text{3.4}}$ \\
HT (pt)~\cite{miech19howto100m} & 8.2 & 24.5 & 35.3 & 24 & - \\
COOT (sc)~\cite{ging2020coot} & 16.7$_{\!\pm\!\text{0.4}}$ & 40.2$_{\!\pm\!\text{0.3}}$ & 52.3$_{\!\pm\!\text{0.5}}$ & 9.0$_{\!\pm\!\text{0.0}}$ & - \\
MMT (pt)~\cite{gabeur2020mmt}$\dagger$ & 17.2$_{\!\pm\!\text{0.4}}$ & 39.5$_{\!\pm\!\text{0.7}}$ & 51.0$_{\!\pm\!\text{0.5}}$ & 10.0$_{\!\pm\!\text{0.0}}$ & 68.2$_{\!\pm\!\text{0.9}}$ \\
Ours (pt p=0.8) & $\textbf{23.2}_{\!\pm\!\text{0.5}}$ & $\textbf{48.0}_{\!\pm\!\text{0.7}}$ & $\textbf{58.6}_{\!\pm\!\text{0.8}}$ & $\textbf{6.0}_{\!\pm\!\text{0.0}}$ & $\textbf{60.4}_{\!\pm\!\text{3.0}}$ \\
\end{tabular}
\end{center}
\end{table}

In Table \ref{table:MSR-VTT}, we compare MSR-VTT results in two different settings: Training from scratch on MSR-VTT (sc) or pre-training on HowTo100M then fine-tuning on MSR-VTT (pt). When training from scratch, our method has a small drop in performance, when compared to MMT~\cite{gabeur2020mmt}. This is likely due to our approach using only 3 modalities instead of 7. Our method's performance is also weaker than a recent approach SSB~\cite{patrick2021supportset} that uses a modified version of MMT. In the HowTo100M pre-training setting however, our modality masking approach outperforms the standard pre-training used in MMT, even if only 12\% of MSR-VTT annotations are related to speech. Our results are competitive wrt SSB. We also show qualitative results of our method on this dataset in Fig.~\ref{fig:qualitative_results}. Note how we perform well in the examples shown in the top two rows -- both the queries refer to the contents of speech. In the second row, while the correct video is retrieved at rank 5, the other videos in the top 5 also describe school systems, demonstrating the difficulty of the dataset where often a caption may be equally relevant to a number of videos.

\begin{table}[h!]
\begin{center}
\caption{Comparison to state of the art on the 1K-A split~\cite{liu2019use} of the MSR-VTT dataset~\cite{xu2016msrvtt}. 
sc: trained from scratch on MSR-VTT. pt: pre-trained on the HowTo100M dataset, then fine-tuned on MSR-VTT.}
\label{table:MSR-VTT}
\scriptsize
\begin{tabular}
{l | c c c c c}
\multicolumn{1}{c}{} & \multicolumn{5}{c}{Text $\implies$ Video} \\
Method & R@1$\uparrow$ & R@5$\uparrow$ & R@10$\uparrow$ & MdR$\downarrow$ & MnR$\downarrow$ \\
\hline
Random & 0.1 & 0.5 & 1.0 & 500.5 & 500.5 \\
JSFusion (sc)~\cite{Yu2018JSFusion} & 10.2 & 31.2 & 43.2 & 13 & -  \\
HT (sc)~\cite{miech19howto100m} & 12.1 & 35.0 & 48.0 & 12 & -  \\
CE (sc)~\cite{liu2019use} & 20.9$_{\!\pm\!\text{1.2}}$ & 48.8$_{\!\pm\!\text{0.6}}$ & 62.4$_{\!\pm\!\text{0.8}}$ & 6.0$_{\!\pm\!\text{0.0}}$ & 28.2$_{\!\pm\!\text{0.8}}$ \\
MMT (sc)~\cite{gabeur2020mmt} & 24.6$_{\!\pm\!\text{0.4}}$ & 54.0$_{\!\pm\!\text{0.2}}$ & 67.1$_{\!\pm\!\text{0.5}}$ & 4.0$_{\!\pm\!\text{0.0}}$ & 26.7$_{\!\pm\!\text{0.9}}$ \\
Ours (sc)& 22.5$_{\!\pm\!\text{0.9}}$ & 53.2$_{\!\pm\!\text{1.5}}$ & 67.1$_{\!\pm\!\text{0.4}}$ & 4.7$_{\!\pm\!\text{0.5}}$ & 25.8$_{\!\pm\!\text{0.3}}$ \\
SSB (sc)~\cite{patrick2021supportset}& 27.4 & 56.3 & 67.7 & 3.0 & - \\
\hline
HT (pt)~\cite{miech19howto100m} & 14.9 & 40.2 & 52.8 & 9 & - \\
Hero (pt)~\cite{li2020hero} & 20.5 & 47.6 & 60.9 & - & - \\
FiT (pt)~\cite{bain2021frozen} & 24.1 & - & 63.9 & 5 & - \\
MMT (pt)~\cite{gabeur2020mmt} & 26.6$_{\!\pm\!\text{1.0}}$ & 57.1$_{\!\pm\!\text{1.0}}$ & 69.6$_{\!\pm\!\text{0.0}}$ & 24.0$_{\!\pm\!\text{0.8}}$ \\
SSB (pt)~\cite{patrick2021supportset}& \textbf{30.1} & 58.5 & 69.3 & \textbf{3.0} & - \\
Ours (pt p=0.8) & 28.7$_{\!\pm\!\text{0.7}}$ & $\textbf{59.5}_{\!\pm\!\text{0.7}}$ & $\textbf{70.3}_{\!\pm\!\text{0.7}}$ & 3.8$_{\!\pm\!\text{0.2}}$ & $\textbf{23.0}_{\!\pm\!\text{0.5}}$ \\
\end{tabular}
\end{center}
\end{table}

Results on ActivityNet are presented in Table \ref{table:ActivityNet}. The annotators of this dataset were explicitly required to ignore the audio track when describing the videos, therefore focusing the descriptions towards the visual modality. Our multi-modal pre-training approach hence yields similar results to the previous state-of-the-art method (SSB~\cite{patrick2021supportset}).

\begin{table}[h!]
\begin{center}
\caption{Paragraph to video retrieval performance on the ActivityNet dataset~\cite{krishna2017activitynet}. sc: trained from scratch on ActivityNet. pt: pre-trained on the HowTo100M dataset, then fine-tuned on ActivityNet.}
\label{table:ActivityNet}
\scriptsize
\begin{tabular}{l | c c c c c}
\multicolumn{1}{c}{} & \multicolumn{5}{c}{Text $\implies$ Video} \\
Method & R@1$\uparrow$ & R@5$\uparrow$ & R@50$\uparrow$ & MdR$\downarrow$ & MnR$\downarrow$ \\
\hline
Random & 0.02 & 0.1 & 1.02 & 2458.5 & 2458.5 \\
FSE (sc)~\cite{zhang2018HSE} & 18.2$_{\!\pm\!\text{0.2}}$ & 44.8$_{\!\pm\!\text{0.4}}$ & 89.1$_{\!\pm\!\text{0.3}}$ & 7 & - \\
CE (sc)~\cite{liu2019use} & 18.2$_{\!\pm\!\text{0.3}}$ & 47.7$_{\!\pm\!\text{0.4}}$ & 6.0$_{\!\pm\!\text{0.0}}$ & 23.1$_{\!\pm\!\text{0.5}}$ \\
HSE (sc)~\cite{zhang2018HSE}& 20.5 & 49.3 & - & - & - \\
MMT (pt)~\cite{gabeur2020mmt} & 28.7$_{\!\pm\!\text{0.2}}$ & 61.4$_{\!\pm\!\text{0.2}}$ & 94.5$_{\!\pm\!\text{0.0}}$ & 3.3$_{\!\pm\!\text{0.5}}$ & $\textbf{16.0}_{\!\pm\!\text{0.4}}$ \\
SSB (pt)~\cite{patrick2021supportset}& \textbf{29.2} & 61.6 & \textbf{94.7} & \textbf{3.0} & - \\
Ours (pt p=0.8) & 29.0$_{\!\pm\!\text{0.5}}$ & $\textbf{61.7}_{\!\pm\!\text{0.3}}$ & 94.6$_{\!\pm\!\text{0.2}}$ & 4.0$_{\!\pm\!\text{0.0}}$ & 16.8$_{\!\pm\!\text{0.5}}$ \\
\end{tabular}
\end{center}
\end{table}

\section{Conclusion}
We present a new pre-training method for learning a multimodal video encoder. It consists of an alternating modality masking strategy, where we mask and predict a different modality at each batch using the other available modalities. We show that this allows us to effectively pre-train a video encoder to jointly process RGB, audio and ASR, even on unlabelled datasets without manually-generated captions. Our method produces competitive results on five downstream video retrieval benchmarks, and is particularly suitable when user queries relate to the spoken language in videos. While our multimodal encoder operates on pre-extracted features, future work will investigate training on raw pixels and audio directly, using newer multimodal transformer architectures\cite{jaegle2021perceiver,nagrani2021attention}.\\

\noindent {\bf Acknowledgements.} This work was supported in part by the ANR
grant AVENUE (ANR-18-CE23-0011).
{\small
\bibliographystyle{ieee_fullname}
\bibliography{lib.bib}
}

\end{document}